\pdfoutput=1

\documentclass[twocolumn]{article}

\usepackage[final]{acl}
\usepackage{enumitem}
\setlist[itemize]{leftmargin=*}
\usepackage{tabularx}
\usepackage{graphicx}
\usepackage{times}
\usepackage{latexsym}
\usepackage{booktabs}
\usepackage{amsmath}
\usepackage{amsfonts}
\usepackage{authblk}   
\usepackage{multirow}

\usepackage[T1]{fontenc}

\usepackage[utf8]{inputenc}
\usepackage[T1]{fontenc}
\usepackage{CJKutf8}

\usepackage{microtype}

\usepackage{inconsolata}

\usepackage{listings}
\usepackage{subcaption}
\usepackage{tablefootnote}
\usepackage[most]{tcolorbox}
\usepackage{fancyhdr}
\usepackage{lipsum}
\usepackage{float}
\usepackage{placeins}

\tcbuselibrary{breakable}

\begin{document}
\begin{CJK}{UTF8}{gbsn}

\title{
SkinFlow: Efficient Information Transmission for Open Dermatological Diagnosis via Dynamic Visual Encoding and Staged RL
}

\author{
  \textbf{Lijun Liu}\textsuperscript{*,1},
  \textbf{Linwei Chen}\textsuperscript{*,1,6,7},
  \textbf{Zhishou Zhang}\textsuperscript{1},
  \textbf{Meng Tian}\textsuperscript{1},
  \\
  \textbf{Hengfu Cui}\textsuperscript{1},
  \textbf{Ruiyang Li}\textsuperscript{1},
  \textbf{Zhaocheng Liu}\textsuperscript{1},
  \textbf{Qiang Ju}\textsuperscript{1},
  \textbf{Qianxi Li}\textsuperscript{\dag,2,3,4,5},
  \textbf{Hong-Yu Zhou}\textsuperscript{\dag,6}
  \\
  \\
  $^1$Baichuan Inc.  \\ 
  $^2$Department of Dermatology, Peking University First Hospital \\
  $^3$Beijing Key Laboratory of Molecular Diagnosis on Dermatoses \\
  $^4$National Clinical Research Center for Skin and Sexually Transmitted Diseases \\
  $^5$NMPA Key Laboratory for Quality Control and Evaluation of Cosmetics \\
  $^6$School of Biomedical Engineering, Tsinghua University \quad $^7$University of Hong Kong \\
  \small{
    \textbf{Correspondence:} \href{hongyu.zhou.ai@gmail.com}{hongyu.zhou.ai@gmail.com}; 
    \href{chancylee7@126.com}{chancylee7@126.com}
  }
  \\
}

\maketitle
\vspace{5em}
\begin{abstract}

General-purpose Large Vision-Language Models (LVLMs), despite their massive scale, often falter in dermatology due to "diffuse attention"—the inability to disentangle subtle pathological lesions from background noise. 
In this paper, we challenge the assumption that parameter scaling is the only path to medical precision. 
We introduce SkinFlow, a framework that treats diagnosis as an optimization of visual information transmission efficiency. 
Our approach utilizes a Virtual-Width Dynamic Vision Encoder (DVE) to "unfold" complex pathological manifolds without physical parameter expansion, coupled with a two-stage Reinforcement Learning strategy. 
This strategy sequentially aligns explicit medical descriptions (Stage I) and reconstructs implicit diagnostic textures (Stage II) within a constrained semantic space. 
Furthermore, we propose a clinically grounded evaluation protocol that prioritizes diagnostic safety and hierarchical relevance over rigid label matching. Empirical results are compelling: our 7B model establishes a new state-of-the-art on the Fitzpatrick17k benchmark, achieving a +12.06\% gain in Top-1 accuracy and a +28.57\% boost in Top-6 accuracy over the massive general-purpose models (e.g., Qwen3VL-235B and GPT-5.2).
These findings demonstrate that optimizing geometric capacity and information flow yields superior diagnostic reasoning compared to raw parameter scaling.

\end{abstract}

\section{Introduction}

Dermatological diagnosis is a visually intensive medical field where diagnostic accuracy relies heavily on the precise identification and interpretation of fine-grained pathological features~\cite{zhang2023recent,badr2025multi}. With the recent advancement of Large Vision-Language Models (LVLMs), there has been significant interest in developing automated systems to assist in skin disease identification\cite{moor2023foundation,zhou2023skingpt,tu2024towards}. However, despite the success of general-purpose models, their application to dermatology remains hindered by two primary challenges. First, general LVLMs often suffer from "diffuse attention," where the model fails to distinguish between critical lesions and irrelevant background noise, leading to suboptimal information transmission. Second, conventional evaluation metrics, such as Top-1 accuracy or exact-match rates, adopt a binary notion of correctness that is fundamentally misaligned with clinical reality. In practice, a diagnosis that captures the correct pathological lineage is far more valuable than a semantically distant misclassification, yet standard metrics treat them as equally incorrect.

In this paper, we address these challenges by reframing dermatological diagnosis as an \textit{information transmission optimization problem}. We conceptualize the model as an \textit{image compression–decoding system} where performance is strictly bounded by the efficiency of information flow: the encoder must compress raw pixels into a high-capacity manifold, while the decoder must reconstruct this data within a constrained diagnostic semantic space. We posit that the failure of existing models stems not from a lack of reasoning power, but from an inability to maximize the "recoverable information" regarding subtle, non-describable pathological cues.

To break the geometric limitations of standard vision backbones, we design the \textbf{Dynamic Visual Encoding (DVE)} module~\cite{chen2025frequency}. As evidenced by our attention attribution analysis, DVE allows the model to "unfold" complex visual manifolds, adaptively suppressing background redundancy and amplifying the signal-to-noise ratio of diagnostic lesions. This architectural innovation enables a transition from uncertain global scanning to high-confidence focal reasoning.

To further maximize this transmission efficiency, we propose a two-stage reinforcement learning framework that systematically decouples visual evidence into \textit{explicit} and \textit{implicit} streams:
\begin{itemize}
    \item \textbf{Stage I (Semantic Alignment via Compression):} We introduce a medical captioning task that forces the model to compress visual information into linguistically interpretable features (describable components), ensuring the retention of explicit clinical signs.
    \item \textbf{Stage II (Diagnostic Refinement via Decoding):} Building on this aligned representation, the model is optimized to reconstruct implicit, non-describable pathological textures within a diagnosis-specific output space, effectively bridging the gap between visual perception and clinical deduction.
\end{itemize}

Furthermore, we challenge the utility of standard metrics in high-stakes medical settings. Moving beyond rigid label matching, we establish a \textbf{Clinically Grounded Evaluation Protocol} inspired by the hierarchical taxonomy of skin diseases~\cite{yan2025derm1m}. This framework prioritizes diagnostic safety and clinical actionability, rewarding therapeutically consistent "near-misses" while strictly penalizing errors that cross critical boundaries (e.g., malignancy).

Empirical results validate our hypothesis that geometric efficiency trumps raw scale. Despite utilizing only 7 billion parameters, our model substantially outperforms general-purpose giants exceeding 200B parameters. On the rigorous \textit{Fitzpatrick17k} benchmark, we establish a new state-of-the-art, achieving a \textbf{+12.06\% gain in Top-1 accuracy} and a massive \textbf{+28.57\% boost in Top-6 accuracy} over the strongest open-source baseline (Qwen3VL-235B), while also surpassing GPT-5.2. These findings suggest that optimizing the compression and decoding of visual information is a more effective mechanism for medical precision than parameter scaling.
On the internal dataset, our model consistently provides a more reliable diagnostic candidate pool, outperforming all competitors in Top-2 through Top-6 metrics (e.g., reaching \textbf{79.21\%} Top-6 accuracy). These findings suggest that enhancing both the compression and decoding efficiency of visual information serves as a fundamental mechanism for improving diagnostic performance and clinical reliability under limited decoding capacity.
Moreover, Our framework naturally supports open-vocabulary dermatological diagnosis, moving beyond fixed-label classification to handle the diverse and long-tailed distribution of real-world skin conditions.

Our main contributions are summarized as follows:
\begin{itemize}
    \item We propose a theoretical framework that treats dermatological VLM training as an optimization of information transmission efficiency, identifying the "recoverable information" bottleneck.
    \item We introduce the Dynamic Visual Encoding (DVE) module, which significantly enhances the visual signal-to-noise ratio by adaptively "unfolding" pathological manifolds.
    \item We implement a two-stage RL training strategy that sequentially masters explicit medical descriptions and implicit diagnostic textures, ensuring robust feature alignment.
    \item We establish a clinical-centric evaluation framework that accounts for disease hierarchies and diagnostic safety, providing a rigorous standard for open-world medical AI.
\end{itemize}

\section{Related Work}

\subsection{Traditional Skin Disease Diagnosis Based on Dermoscopic Data}
Early studies on automated skin disease diagnosis mainly relied on dermoscopic images, which are captured by professional devices\cite{pham2021ai,shahin2018deep,salamaa2021deep,ashraf2022melanoma}. Pham et al \cite{pham2021ai} employed deep convolutional neural networks (CNNs) to achieve high-accuracy melanoma recognition from dermoscopic images. Shahin et al\cite{shahin2018deep} proposed an ensemble model by integrating ResNet-50 and Inception V3, which demonstrated strong performance in distinguishing malignant from benign lesions. Other works combined VGG16 with support vector machines (SVMs) for binary lesion classification\cite{salamaa2021deep}. However, these methods are constrained by their reliance on specialized imaging equipment and the limited availability of annotated datasets, making them difficult to deploy in large-scale health monitoring or telemedicine applications.

\subsection{Single-Modality Visual Recognition Methods}
Traditional deep learning approaches have primarily focused on single modality visual tasks such as lesion classification and segmentation\cite{ashraf2022melanoma,srinivasan2025capsule}. For instance, U-Net and its variants (e.g., ResUNet++) have been widely used for automated lesion segmentation to assist clinicians in delineating lesion boundaries\cite{ashraf2022melanoma,jojoa2021melanoma,araujo2022automatic}. Capsule networks (CapsNet) have also been explored to enhance spatial hierarchical representation, enabling multi-class skin disease recognition and fine-grained classification\cite{srinivasan2025capsule,nawaz2025skin,eskandari2024efficient}. Despite their success, these models typically output static classification results without natural language explanations or interactive reasoning. The lack of interpretability and user interaction makes such “black-box” systems less suitable for clinical practice, where transparent decision-making and physician–patient communication are essential.

\subsection{Multimodal Models in Medical Diagnosis}
With the rapid development of large-scale multimodal models, neural systems are now capable of jointly understanding visual and textual inputs, producing analysis results in natural language form\cite{guo2024llava,el2024democratizing,zhou2023skingpt,yan2025multimodal}. LLaVA-Ultra\cite{guo2024llava} introduced fine-grained visual–language fusion for Chinese ultrasound image question answering, improving medical semantic comprehension. TinyLLaVA-Med\cite{el2024democratizing} achieved efficient inference in low-resource medical settings through lightweight fine-tuning. SkinGPT-4\cite{zhou2023skingpt}, one of the few multimodal models specifically designed for dermatology, adopted a two-stage training process to enable both lesion description generation and diagnosis assistance. Although these methods reveal the promise of multimodal medical diagnosis, comprehensive and systematic research in this domain remains limited.

\subsection{Emergence of Next-Generation Multimodal Foundation Models}
Recent multimodal foundation models, such as Qwen2.5-VL\cite{bai2025qwen2}, Qwen3-VL\cite{qwen3technicalreport}, and InternVL-3\cite{zhu2025internvl3}, have demonstrated outstanding performance in vision language understanding and visual question answering tasks. Domain-specific medical multimodal models, such as Lingshu-32B\cite{xu2025lingshu}, have also shown remarkable progress in medical image interpretation. Nevertheless, how to effectively integrate general-purpose multimodal foundation models into medical applications—and optimize them for domain-specific semantics and diagnostic reasoning—remains an open and underexplored problem.

\subsection{Reinforcement Learning and Generalization}
While supervised fine-tuning (SFT) remains the mainstream post-training paradigm, its generalization ability is inherently limited. Recent studies suggest that SFT can be viewed as an implicit form of policy gradient with hidden reward bias, which may lead to overfitting on narrow data distributions\cite{wu2025generalization}. In contrast, reinforcement learning (RL) explicitly optimizes policies based on reward signals, enabling adaptive improvement in complex task spaces and yielding stronger robustness and generalization\cite{chu2025sft}. However, RL in high-dimensional medical domains still faces challenges in exploration efficiency and training stability\cite{dulac2019challenges,al2024reinforcement}. To address these issues, our work introduces a two-stage reinforcement learning framework that integrates multimodal description learning with reward-driven diagnostic policy optimization, achieving improved accuracy while maintaining stability and scalability.

\section{Method} 
\begin{figure}[t]
  \includegraphics[width=\columnwidth]{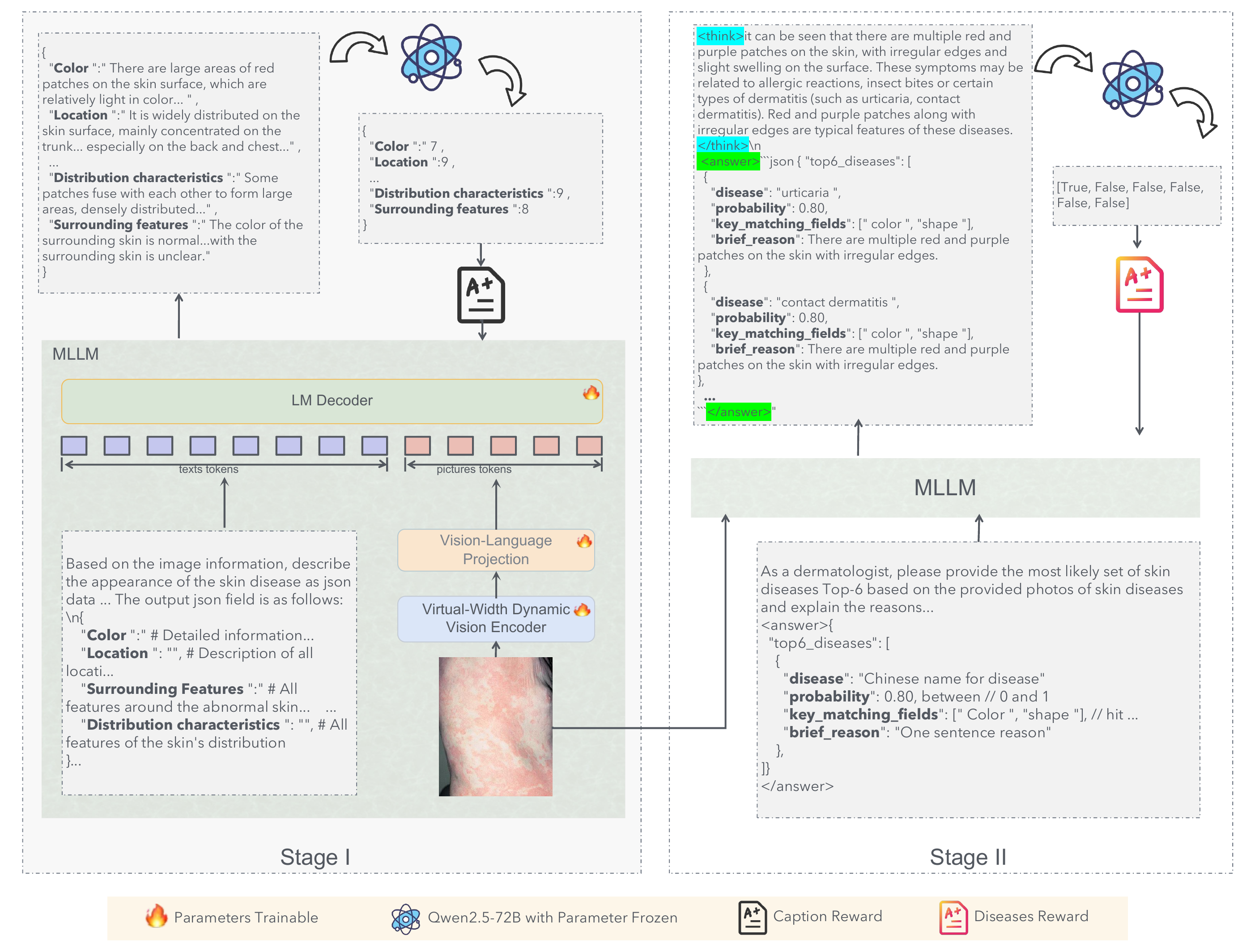}
  \caption{\textbf{Two-stage reinforcement learning framework for dermatological diagnosis}.
    In Stage 1, the model performs medical caption generation. The LLM scores each attribute field of the generated description, and these field-wise scores are integrated into a caption reward to refine medical feature learning. In Stage 2, the model predicts disease categories based on learned representations. The LLM evaluates each prediction, and a customized reward function converts these evaluations into a final diagnostic reward that optimizes classification accuracy and ranking consistency.}
  \label{fig:overview}
\end{figure}
This section details the proposed two-stage training pipeline for dermatological diagnosis (Figure~\ref{fig:overview}). We begin by formalizing our framework from an \textit{information transmission} perspective, providing the theoretical rationale that unifies the staged design. Building upon this foundation, we introduce the Dynamic Vision Encoder (DVE) module, which addresses the visual representation bottlenecks in current multimodal LLMs and the RL algorithm used for optimization. Finally, we provide a detailed implementation of Stage I (caption learning) and Stage II (diagnosis learning), elaborating on the construction of dermatological datasets and the design of our clinically-grounded reward functions.

\subsection{Overall Framework: An Information Transmission Perspective}

To optimize the diagnostic performance of our model, we formalize the training process as an \textit{image compression–decoding} task. As illustrated in our framework, the model's efficacy is determined by the information transmission efficiency from the raw pixels to the final semantic diagnosis. We define the total visual information $\mathcal{I}$ as comprising two distinct components: describable features $\mathcal{I}_d$ (explicit medical signs) and non-describable features $\mathcal{I}_n$ (implicit pathological textures).

The core rationale for our two-stage strategy is to maximize the \textit{recoverable information} within a constrained decoding space. 
\begin{itemize}
    \item \textbf{Stage I (Information Compression):} By introducing the medical captioning task, we force the encoder to prioritize the compression of $\mathcal{I}_d$ into linguistically interpretable representations. This stage establishes a high-capacity channel for key diagnostic features.
    \item \textbf{Stage II (Semantic Decoding):} Upon the foundation of high-quality representations, the model is then fine-tuned to integrate $\mathcal{I}_n$ and decode the joint information into diagnosis-specific semantics. 
\end{itemize}
By optimizing these processes in tandem, the model's information transmission approaches the theoretical upper bound of the diagnostic space, ensuring that the final output is grounded in both explicit clinical evidence and implicit visual cues.

\subsection{Virtual-Width Dynamic Vision Encoder}
\label{sec:vision_encoder}

While existing Multimodal LLMs (MLLMs) achieve great succuss, a stark asymmetry exists in current MLLMs: while the LLM backbones have scaled to billions of parameters to support advanced cognitive reasoning, the vision encoders remain disproportionately lightweight (e.g., vision encoder $\sim$0.6B vs.\ 7B LLM in Qwen2.5-VL). This imposes a fundamental \textit{representation bottleneck}: the model possesses a powerful "brain" for semantic processing but is limited by a "retina" with insufficient geometric capacity, restricting its ability to discern subtle pathological features essential for diagnosis.

To address this asymmetry without incurring the massive computational cost of scaling the vision encoder, we introduce {FDLinear (Frequency Dynamic Linear)}~\cite{chen2025frequency}, a parameter-efficient dynamic operator. By replacing static linear layers with FDLinear in MLPs, we exponentially expand the effective geometric capacity of the vision encoder.

\subsubsection{Theoretical Motivation: Escaping the Capacity Curse of Cover's Theorem}
Standard Vision Transformers rely on static linear layers for feature mixing. Formally, given an input feature $x \in \mathbb{R}^d$, a static layer computes $y = Wx + b$, where $W \in \mathbb{R}^{d \times d}$ is fixed after training.

According to {Cover's Theorem}~\cite{cover2006geometrical} on the geometrical separation of patterns, the probability $P(N, d)$ that $N$ random patterns are linearly separable in a $d$-dimensional space is:
\begin{equation}
    P(N, d) \approx \begin{cases} 
    1 & \text{if } N \le 2d \\
    0 & \text{if } N \gg 2d 
    \end{cases}
\end{equation}
In complex dermatological diagnosis, the number of visual patterns (texture, erythema, scale, ulcers) $N$ can be very large ($\to \infty$), while the physical dimension $d$ of the vision encoder is fixed (e.g., $d=1280$). Consequently, a static encoder suffers from {\it Capacity Collapse}, forcing the model to average out fine-grained details to satisfy the global optimization objective.

\begin{figure}[t]
  \centering
  \includegraphics[width=0.68\linewidth]{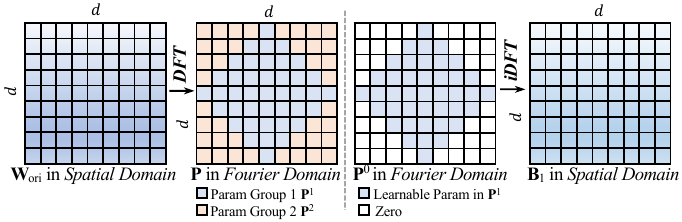} 
  \caption{\textbf{Illustration of frequency disjoint basis construction.} 
  The process transforms the weight matrix $\mathbf{W}_{ori} \in \mathbb{R}^{d\times d}$ from the spatial domain to the Fourier domain via the Discrete Fourier Transform (DFT). The frequency spectrum is then partitioned into disjoint groups based on frequency index (e.g., $\mathbf{P}^1$ corresponds to the central low-frequency components in blue, while $P^2$ corresponds to the peripheral high-frequency components in orange).
  To generate a specific spatial basis $\mathbf{B}_1$, we retain only the learnable parameters belonging to Group 1 ($\mathbf{P}^1$) and mask all other frequency indices to zero. Finally, an inverse DFT (iDFT) reconstructs the spatial basis matrix. This design ensures that each basis $\mathbf{B}_k$ specializes in a distinct frequency band, minimizing spectral redundancy.}
  \label{fig:fdw_construction}
\end{figure}

\subsubsection{FDLinear: Implicit High-Dimensional Mapping}

A straightforward approach to satisfy the geometric separability condition ($N \le 2d$) would be to physically expand the channel dimension $d$. However, this strategy is computationally intractable. Increasing channel width leads to a quadratic growth in parameters ($O(d^2)$). Furthermore, due to the inherent 2D spatial nature of images, increasing resolution to capture fine pathological details results in a \textit{quadratic explosion} in the number of visual tokens. Coupling a massive physical dimension with this surge in token count would cause computational complexity (FLOPs) to skyrocket, rendering the model undeployable in real-world clinical settings.

We propose a paradigm shift from \textit{physical dimension expansion} to \textit{virtual dimension expansion}. FDLinear circumvents the computational wall by decoupling the weight space into $K$ orthogonal spectral bases $\{\mathbf{B}_1, \dots, \mathbf{B}_K\}$. 
As illustrated in Figure~\ref{fig:fdw_construction}, these bases are not randomly initialized but are constructed via {Frequency Disjoint Partitioning}. We partition the full frequency spectrum of the weight matrix into $K$ disjoint groups (e.g., concentric frequency bands). Each basis $\mathbf{B}_k$ is generated by retaining only the parameters in the $k$-th frequency group and applying an inverse Discrete Fourier Transform (iDFT). 
The dynamic weight $W(\bar x)$ is constructed via a context-aware linear combination:
\begin{equation}
    W(\bar x) = \sum_{k=1}^{K} \alpha_k(\bar x) \cdot \mathbf{B}_k
\end{equation}
Crucially, to maintain extreme parameter efficiency, the dynamic coefficient $\alpha_k(\bar x)$ is not a dense matrix but is factorized into three orthogonal modulation vectors (corresponding to the input dimension, output dimension, and basis dimension $K$), all of which are predicted by a single, highly lightweight fully-connected (FC) bottleneck layer. $\bar x$ is image-level global average result.

\vspace{0.5ex}
\noindent\textbf{The Virtual Dimension Expansion Mechanism:} \\
Mathematically, the operation $y = W(\bar x)x$ can be viewed as a two-step "Expand-and-Collapse" process. If we were to explicitly compute the projection of $x$ onto all bases, we would generate a massive hidden representation $\mathcal{H} \in \mathbb{R}^{K \times d}$:
\begin{equation}
    \mathcal{H} = \text{Concat}(\mathbf{B}_1 x, \mathbf{B}_2 x, \dots, \mathbf{B}_K x)
\end{equation}
Here, the channel dimension virtually explodes from $d$ to $K \times d$ (e.g., $1280 \to 81,920$ for $K=64$). In this hyperspace, the complex visual manifolds are "unfolded" and become linearly separable according to Cover's Theorem~\cite{cover2006geometrical}.

\vspace{0.5ex}
\noindent\textbf{Computational Efficiency Implementation:} \\
Crucially, {we never explicitly materialize this massive vector $\mathcal{H}$}. By exploiting the linearity of matrix operations, we fuse the aggregation step (the weighted sum) \textit{before} the matrix multiplication:
\begin{equation}
    y = \underbrace{\left( \sum_{k=1}^{K} \alpha_k(\bar x) \cdot \mathbf{B}_k \right)}_{\text{Pre-computed } W(\bar x) \in \mathbb{R}^{d \times d}} \cdot x
\end{equation}
Instead of performing calculations in the exploding $K \times d$ dimension (which would require $O(K \cdot d^2)$ FLOPs per token), we essentially "compress" the $K$ weight basis into a single dynamic matrix $W(\bar x)$ of size $d \times d$. This allows the model to enjoy the \textit{geometric capacity} of a massive width ($K \times d$) while maintaining the \textit{computational footprint} of a compact layer ($d$).

\begin{figure*}[t]
  \centering
  \includegraphics[width=\linewidth]{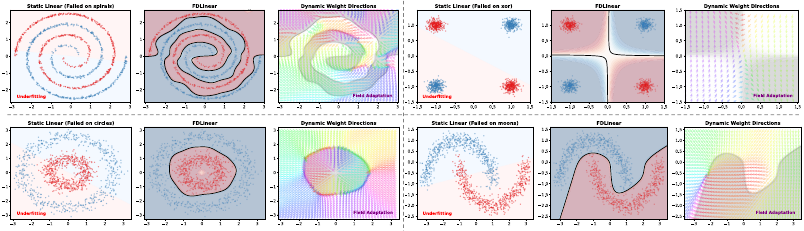} 
  \caption{\textbf{Visualizing manifold unfolding and virtual capacity.} 
  We evaluate the geometric representation capability on four classic non-linearly separable datasets: Spirals, XOR, Circles, and Moons.
  {(Left column of each sub-figure)} Constrained by Cover's Theorem, a standard static layer (width $d=2$) is topologically restricted to a single separating hyperplane, resulting in severe underfitting (accuracy $\approx 50\%$).
  {(Middle column of each sub-figure)} Without increasing the physical width ($d=2$), FDLinear ($K=12$) dynamically constructs high-order decision boundaries, successfully disentangling complex manifolds (e.g., the intertwined spirals). This empirically validates the "Virtual Width Expansion" hypothesis.
  {(Right column of each sub-figure)} The vector fields visualize the orientation of the generated weight matrix $W(\bar x)$ across the input space. The rotating and radiating patterns (labeled "Field Adaptation") demonstrate that FDLinear acts as a \textit{sample adaptive layer}, modulating its projection direction based on local geometric curvature.}
  \label{fig:toy_vis}
\end{figure*}

\subsubsection{Visual Verification and Efficiency Analysis}
\label{sec:efficiency}

To empirically validate the "virtual width expansion" hypothesis, we visualized the decision boundaries of FDLinear on classic non-linearly separable manifolds (Figure~\ref{fig:toy_vis}). 
As predicted by Cover's Theorem, the static linear baseline suffers from \textit{capacity collapse}, failing to separate complex topologies like intertwined spirals or concentric circles. In contrast, FDLinear successfully "unfolds" these manifolds. 
Crucially, the {dynamic weight directions} (Figure~\ref{fig:toy_vis}, right column of each sub-figure) reveal a non-uniform, rotating vector field. This confirms that the model is not merely memorizing data, but is actively modulating its "gaze" (projection direction) according to the local geometry of the input features.

This geometric adaptability translates into a paradigm shift in efficient scaling for medical AI:

\begin{itemize}[leftmargin=2.5em, itemsep=0pt, parsep=0pt, topsep=-\parskip]
    \item \textbf{Geometric Capacity beyond Physical Limits:} As evidenced by the "spirals" experiment, FDLinear enables a compact layer to model decision boundaries that typically require vast parameter expansion. This effectively bridges the capacity gap between the 0.6B vision perceiver and the 7B language reasoner.
    
    \item \textbf{Parameter-Efficient Storage:} Despite the virtual width expansion, we store only the bases $\mathbf{B}$ and a lightweight coefficient predictor. The parameter overhead is negligible ($<5\%$) compared to the dense matrices required for physical scaling.
    
    
    \item \textbf{Inference Speed:} Since the dynamic composition $W(x)$ collapses into a standard linear transformation at the operator level, the computational complexity remains comparable to standard linear layer, avoiding the high latency and memory fragmentation associated with MoE (Mixture of Experts) architectures.
\end{itemize}

By "virtualizing" the width of the vision encoder, SkinFlow achieves high-fidelity diagnostic reasoning at a fraction of the cost of training a multi-billion parameter vision backbone.

\subsection{Preliminary: RL algorithm}
In contrast to approaches that explicitly replicate intermediate reasoning steps, RLVR \cite{guo2025deepseek,team2025kimi} relies solely on outcome-driven feedback, facilitating scalable reinforcement learning across extensive task datasets.

Group Relative Policy Optimization (GRPO) \cite{guo2025deepseek} is an efficient RL algorithm that eliminates the need for a separate critic model.
Given a query $q$,We adopt Group Relative Policy Optimization (GRPO) \cite{guo2025deepseek} as our RL backbone. GRPO is a sample-efficient variant of policy optimization that eliminates the need for a separate critic by operating on groups of outputs. Given a query $q$, GRPO samples a group of $G$ candidate outputs ${o_1,\dots,o_G}$ from the current policy $\pi_{\theta_{\text{old}}}$ and evaluates them using a task-specific reward function to yield rewards ${r_1,\dots,r_G}$. Advantages are computed by normalizing within the sampled group:
\begin{equation}
A_i = \frac{r_i - \operatorname{mean}{r_j \mid j=1}^G)}{\operatorname{std}({r_j\mid}{j=1}^G)}.
\end{equation}
The GRPO objective optimizes a clipped importance-weighted surrogate with a KL penalty to a reference policy:
\begin{align}
\mathcal{J}{\text{GRPO}}(\theta) &= \mathbb{E}{q \sim \mathcal{D}, {o_i \mid i=1}^G \sim \pi{\theta_{\text{old}}}} \frac{1}{G} \sum_{i=1}^{G} \frac{1}{|o_i|} \sum_{t=1}^{|o_i|} \Big(
\min \big( \varphi_{i,t}(\theta) A_{i,t},\ \mathrm{clip}(\varphi_{i,t}(\theta),1-\epsilon,1+\epsilon)A_{i,t} \big) \notag\
&\qquad\qquad\qquad\qquad - \beta, \mathbb{D}{\mathrm{KL}}(\pi\theta|\pi_{\mathrm{ref}}) \Big),
\end{align}
where $\varphi_{i,t}(\theta)=\dfrac{\pi_\theta(o_{i,t}\mid q,o_{i,<t})}{\pi_{\theta_{\text{old}}}(o_{i,t}\mid q,o_{i,<t})}$. GRPO’s group normalization stabilizes training and enables efficient utilization of reward signals from diverse candidate outputs.

\subsection{Stage I: Learning Dermatological Image Descriptions}

The primary goal of this stage is to enable the model to produce accurate and clinically coherent descriptions for dermatological images.
Compared with supervised fine-tuning (SFT), reinforcement learning mitigates the problem of entropy collapse and achieves better generalization, especially under limited annotation conditions.

\subsubsection{Construction of Dermatological Description Samples}

We collected approximately 5,000 dermatological images from both domestic and international sources, each annotated with disease category labels.
To obtain high-quality description samples (captions), we employed a hybrid strategy combining large language model (LLM)-based generation and expert refinement.
Among them, 4,000 captions were automatically annotated through the machine labeling process described below.

\vspace{1ex} 
\noindent\textbf{Machine annotation process:}

1. Use a multimodal large language model (MLLM) to generate an initial caption.

To facilitate the subsequent quantitative evaluation of caption quality, we guided the model to produce structured medical captions. The process for determining the structured fields is as follows:
    \begin{itemize}[leftmargin=2.5em, itemsep=0pt, parsep=0pt, topsep=-\parskip]
        \item Generation of unstructured captions: The large language model (LLM) first generates rich, descriptive free-text captions of skin lesion appearances.
        \item Structuring of captions: The unstructured captions are then reformatted into a structured form through LLM-based processing.
        \item Field frequency analysis: The frequency of each field is calculated and ranked.
        \item Manual refinement: Based on the ranked results, experts manually determine the final set of structured fields and corresponding annotation guidelines.
    \end{itemize}

2. Input the generated caption into an LLM to infer the corresponding diagnosis.

3. Compare the inferred diagnosis with the ground-truth disease label.

    \begin{itemize}[leftmargin=2.5em, itemsep=0pt, parsep=0pt, topsep=-\parskip]
        \item If consistent, the caption is accepted.
        \item If inconsistent, regenerate the caption and repeat Steps (2)–(3) up to five times.
    \end{itemize}
    
4. If a consistent caption is not obtained after five iterations, the sample is passed to human experts for revision.

\vspace{1ex} 
\noindent\textbf{Human annotation process:}

1. Use an MLLM to generate an initial caption.

2. Medical experts manually revise the caption based on professional knowledge.

3. The revised caption is re-evaluated by an LLM for diagnostic consistency.

4. If the LLM-determined diagnostic accuracy meets the quality threshold, the caption is accepted; otherwise, the revision process is repeated.

\subsubsection{Reward Function Design}

To ensure that the generated captions are both clinically valid and semantically complete, we designed a multi-dimensional reward mechanism.For attributes involving continuous degrees of variation, such as lesion color and size, we formulated instruction-based rules to measure the correlation between the predicted and ground-truth descriptions. For attributes with well-defined medical categories, such as lesion type, we constrained the model’s outputs through instructional rules that require selection within a predefined set of medically valid options.
For each predefined attribute, the LLM assigns a score ranging from 0 to 10, with scores ≥6 considered acceptable.
The overall reward is computed as a weighted average of individual attribute scores:
\\
\begin{equation}
R = \sum_i \alpha_i \cdot s_i
\end{equation}
\\
where ( $s_i$ ) denotes the attribute score and ( $\alpha_i$ ) represents its corresponding weight.

To validate the reliability of the reward design, we calculated the correlation between the generated caption reward and the diagnostic accuracy obtained when using the caption as input for inference.
A strong positive correlation indicates that the designed reward provides an effective optimization signal.
Table \ref{tab:reward} presents the average rewards and diagnostic accuracies under the optimal scoring configuration.

Determination of attribute weights
The weight of each attribute (( $\alpha_i$ )) is determined according to its frequency of use in the LLM’s diagnostic reasoning process.
Attributes that are more frequently referenced as diagnostic evidence are assigned higher weights accordingly.

\begin{table}[htbp]
\centering
\caption{Correlation between caption reward and diagnostic accuracy across different multimodal models.}
\label{tab:reward}
\begin{tabular}{lcc}
\toprule
\textbf{Model Name} & \textbf{Caption Reward} & \textbf{Diagnostic Accuracy} \\
\midrule
Qwen2.5-VL-7B & 6.162 & 13.79\% \\
Qwen2.5-VL-72B & 6.688 & 23.00\% \\
Lingshu-32B & 5.924 & 24.14\% \\
InternVL3-78B & 6.912 & 26.44\% \\
Qwen3-VL-Instruct-235B-A22B & 7.186 & 27.59\% \\
\bottomrule
\end{tabular}
\end{table}

\subsection{Stage II: Dermatological Diagnosis Training}

In the second stage, the model aims to predict the top-K most probable diagnoses for a given dermatological image.
Training is again conducted under an RL framework rather than SFT, for two key reasons:

1. Terminological diversity: A single disease often has multiple equivalent medical names.The rigid token-level matching of SFT is thus unsuitable for learning semantically equivalent expressions.

2. Efficiency in top-K learning: SFT requires explicit construction of all possible top-K labels, whereas RL flexibly assigns rewards without enumerating the entire label space.

\vspace{1ex} 
\noindent\textbf{Reward Function Design}

The model outputs a ranked list of top-K candidate diagnoses.
Let the position weights be ( $w = [w_1, w_2, ..., w_K] $), and let the correctness list be ( $l = [l_1, l_2, ..., l_K]$ ), where ( $l_i \in {0, 1}$ ) is determined by an LLM-based verification process.

If the correct diagnosis first appears at position ( $i$ ), the final reward is defined as the corresponding positional weight ( $w_i$ ):
\begin{equation}
i = \min \{ k \in \mathbb{N} \mid 1 \le k \le K, l_k = 1 \}
\end{equation}
\begin{equation}
r = w_i
\end{equation}
This reward design ensures that the model is rewarded for correctness while also being encouraged to rank the correct diagnosis as high as possible in its prediction list.

\section{Experiments}
\subsection{Experimental Setup}

\vspace{1ex} 
\subsubsection{Implementation Details}

\textbf{Model Architecture:} We initialized our model based on Qwen2.5-VL-Instruct-7B~\cite{bai2025qwen2}. To construct the \textit{Virtual-Width Dynamic Vision Encoder}, we replaced the static linear layers within the MLPs of the original Vision Transformer with our proposed FDLinear operators at layers 8, 16, 24 and 32. We set the number of spectral bases $d/2$, where $d$ is the input dimension. effectively expanding the virtual geometric width by a large factor. Despite this substantial capacity boost, the additional parameter overhead for storing bases and coefficient predictors is less than 5\% of the original vision encoder size.

\textbf{Training Protocol:} 
The first-stage training was conducted based on the Qwen2.5-VL-Instruct-7B model\cite{bai2025qwen2}. The AdamW optimizer was employed with a learning rate of ($1\times10^{-6}$), and a cosine warmup strategy was used for scheduling.
The second-stage training continued from the first-stage checkpoint under a reinforcement learning (RL) framework, with a learning rate of ($5\times10^{-7}$).
The entire RL pipeline was implemented using the VERL framework \cite{sheng2024hybridflow}.

\vspace{1ex} 
\subsubsection{Test Dataset}

To evaluate dermatological diagnostic performance under realistic clinical settings, we construct two evaluation benchmarks that do not assume a closed-set disease taxonomy. Together, the two benchmarks cover approximately 200 distinct skin disease categories, reflecting the diversity and long-tailed nature of real-world dermatological practice.

The first benchmark is derived from the publicly available Fitzpatrick17k dataset. From this dataset, we randomly sample 1,000 images spanning a broad range of dermatological conditions. As a widely used open-source benchmark, Fitzpatrick17k provides a representative and diverse test bed for assessing generalization performance.

The second benchmark is an internally curated dataset consisting of approximately 200 images. All samples in this internal dataset were independently reviewed and corrected by board-certified dermatologists from Class-III Grade-A hospitals, each with more than five years of clinical experience. This rigorous expert verification process ensures high diagnostic accuracy of the ground-truth labels and enhances the objectivity and reliability of the evaluation.

Importantly, neither benchmark restricts the diagnosis space to a fixed or exhaustive disease list. Instead, the evaluation setting allows for hierarchical and semantically related diagnoses, which more closely mirrors real-world clinical scenarios where disease boundaries may overlap and exact subtype distinctions are not always required for effective clinical decision-making.

\vspace{1ex} 
\subsubsection{Baselines}

We compared our proposed SkinFlow model with several representative multimodal large language models (MLLMs), including:
\begin{itemize}[leftmargin=2.5em, itemsep=0pt, parsep=0pt, topsep=-\parskip]
    \item General-purpose MLLMs:
  Qwen2.5-VL-Instruct-7B\cite{bai2025qwen2}, InternVL3-78B\cite{zhu2025internvl3}, Qwen3-VL-Instruct-235B-A22B\cite{qwen3technicalreport} and GPT-5.2\cite{openai2025gpt52}.
   \item Medical-domain MLLM:Lingshu-32B \cite{xu2025lingshu},medgemma-27b-it \cite{sellergren2025medgemma} .
\end{itemize}

\vspace{1ex} 
\subsubsection{Evaluation Metrics}

Conventional evaluation metrics for classification tasks, such as accuracy or exact-match rate, adopt a binary notion of correctness, where any prediction that does not exactly match the reference label is treated as equally incorrect. However, this evaluation paradigm is fundamentally misaligned with real-world clinical practice, particularly in dermatological diagnosis. In clinical settings, diagnostic predictions often exhibit varying degrees of semantic proximity and therapeutic relevance. A diagnosis that is not strictly identical to the reference label may still provide substantial clinical value if it leads to an appropriate treatment plan, whereas a semantically distant but technically distinct diagnosis may result in harmful clinical decisions.

Recent studies have emphasized that medical AI systems should be evaluated based on their ability to support clinical reasoning and decision-making rather than mere label correspondence. For example, \cite{sokol2025artificial} argue that AI systems should genuinely enhance clinical decision quality to bridge the translational gap between algorithmic performance and real-world utility. In the dermatology domain, Derm1M \cite{yan2025derm1m} further highlights the strong hierarchical structure among skin diseases, where diagnoses along the same pathological lineage (e.g., parent–child relations) are often more clinically informative than unrelated categories.

Motivated by these insights, we adopt a clinically grounded evaluation protocol instead of conventional accuracy. Our evaluation explicitly accounts for the hierarchical relationships between dermatological diseases and their corresponding clinical implications, particularly treatment consistency and diagnostic safety. Specifically, model predictions are categorized according to the following criteria:
\begin{itemize}[leftmargin=2.5em, itemsep=0pt, parsep=0pt, topsep=-\parskip]
    \item True (Correct Diagnosis):
The predicted diagnosis exactly matches the reference label or corresponds to a medically accepted synonym, alias, or abbreviation (e.g., Herpes zoster ≡ Shingles).

    \item True (Subclass Match):
The predicted diagnosis is a clinically valid subclass of the reference diagnosis (e.g., Atopic dermatitis → Eczema), reflecting higher diagnostic specificity while remaining therapeutically consistent.

    \item Parent-Class Predictions:
    \begin{itemize}[leftmargin=2.5em, itemsep=0pt, parsep=0pt, topsep=-\parskip]
        \item True: The predicted diagnosis is a closely related parent category that retains clear clinical value, where treatment strategies are consistent or differ only in lesion location.

        \item True (Coarse but Directionally Correct): The prediction captures the correct diagnostic direction but lacks specificity; nevertheless, it remains clinically actionable.

        \item False: The predicted parent category is overly broad and fails to provide meaningful guidance for clinical decision-making (e.g., Dermatitis, Skin cancer).
    \end{itemize}
    \item False (Sibling-Class Confusion):
The predicted diagnosis belongs to a sibling category of the reference disease, representing a common but clinically misleading misclassification.

    \item False (Safety-Critical Errors):
Predictions that cross critical clinical boundaries, such as benign vs. malignant or infectious vs. non-infectious diseases, are strictly penalized in accordance with the medical principle of “First, do no harm.”

    \item False (Invalid or Irrelevant Predictions):
The predicted diagnosis is empty or exhibits no meaningful clinical relationship to the reference label.
\end{itemize}
Under this evaluation framework, diagnostic predictions are judged by their clinical actionability and safety, rather than by rigid label equivalence. This design allows us to distinguish between clinically meaningful near-miss predictions and dangerous or non-informative errors, thereby providing a more realistic and clinically relevant assessment of model performance in open-world dermatological diagnosis scenarios.

All models were evaluated using identical prompts for fair comparison.
the evaluation was conducted using Gemini-2.5-Pro.
Each evaluation was repeated three times and the mean accuracy was reported to mitigate randomness.

\subsection{Main Results}

Table \ref{tab:comparison} presents a comprehensive quantitative comparison across Fitzpatrick17k and the Self-owned datasets. Our 7B model achieves state-of-the-art (SOTA) performance on nearly all benchmarks, consistently outperforming both massive general-purpose VLMs and specialized medical models.
\noindent\textbf{Superiority on Public Benchmarks.}
The most striking results are observed on the \textit{Fitzpatrick17k} dataset. While general VLMs and specialized medical models struggle with the fine-grained diversity of skin diseases, our model achieves a \textbf{Top-1 accuracy of 29.19\%}, surpassing the strongest baseline (GPT-5.2) by \textbf{10.95\%} and the massive Qwen3VL-235B by \textbf{12.06\%}. Furthermore, our model exhibits exceptional retrieval capability in complex cases, reaching a \textbf{Top-6 accuracy of 71.16\%}, which represents a \textbf{+28.57\%} boost over the Qwen3VL-235B baseline. This substantial margin validates the effectiveness of our two-stage strategy in mastering intricate pathological patterns that general models fail to capture.
\noindent\textbf{Parameter Efficiency vs. Massive Models.}
Despite being orders of magnitude smaller (7B parameters), our model demonstrates superior performance and efficiency compared to "super-large" models. On the \textit{Self-owned dataset}, although GPT-5.2 maintains a slight edge in Top-1 accuracy (39.11\% vs. 36.63\%), our model surpasses it in all other ranking metrics (\textit{Top-2 to Top-6}). Notably, our model achieves a \textbf{79.21\% Top-6 accuracy} on the Self-owned set, significantly higher than GPT-5.2 (68.81\%) and Qwen3VL-235B (64.00\%). This suggests that while general models may identify the primary label in familiar contexts, our model provides a much more robust and clinically relevant diagnostic "candidate pool," which is critical for reducing omissions in real-world dermatology.

\begin{table*}[t]
\centering
\caption{Comparison of different models on Self-owned dataset and Fitzpatrick17k.}
\label{tab:comparison}
\small
\setlength{\tabcolsep}{3.5pt} 
\resizebox{\textwidth}{!}{
\begin{tabular}{ll c cccccc cccccc}
\toprule
\multirow{2}{*}{\textbf{Domain}} & \multirow{2}{*}{\textbf{Model}} & \multirow{2}{*}{\textbf{\shortstack{Param\\Size}}} & \multicolumn{6}{c}{\textbf{Fitzpatrick17k}} & \multicolumn{6}{c}{\textbf{Self-owned dataset}} \\
\cmidrule(lr){4-9} \cmidrule(lr){10-15}
& & & TOP1 & TOP2 & TOP3 & TOP4 & TOP5 & TOP6 & TOP1 & TOP2 & TOP3 & TOP4 & TOP5 & TOP6 \\
\midrule
\multirow{6}{*}{General} & Qwen2.5VL-7B-Instruct & 7B & 10.05 & 17.13 & 21.34 & 24.59 & 27.66 & 31.00 & 23.98 & 37.24 & 43.88 & 45.92 & 50.51 & 54.59 \\
& Qwen2.5VL-72B-Instruct & 72B & 11.75 & 19.01 & 23.59 & 27.41 & 31.04 & 33.62 & 26.73 & 42.57 & 51.98 & 56.93 & 60.40 & 64.85 \\
& Qwen3VL-32B-Instruct & 32B & 16.02 & 23.72 & 28.03 & 32.33 & 36.74 & 40.34 & 30.65 & 47.24 & 50.25 & 55.28 & 59.80 & 64.32 \\
& Qwen3VL-235B-A22B-Instruct & 235B & 17.13 & \underline{25.75} & \underline{31.96} & 35.57 & \underline{39.78} & 42.59 & 35.43 & \underline{50.29} & 56.00 & 57.71 & 60.57 & 64.00 \\
& Internvl3-78B & 78B & 12.70 & 22.16 & 27.79 & 32.76 & 35.82 & 38.30 & 28.22 & 46.04 & 53.47 & 56.93 & 60.40 & 62.87 \\
& GPT-5.2 & / & \underline{18.24} & 25.69 & 31.52 & \underline{36.49} & 39.16 & \underline{42.88} & \textbf{39.11} & 48.51 & 56.93 & \underline{63.86} & \underline{66.34} & \underline{68.81} \\
\midrule
\multirow{2}{*}{Medical} & medgemma-27b-it & 27B & 13.60 & 21.23 & 26.13 & 30.72 & 33.95 & 37.87 & 22.87 & 28.72 & 34.04 & 45.74 & 51.06 & 54.79 \\
& Lingshu-32B & 32B & 10.06 & 17.91 & 23.28 & 26.63 & 29.69 & 32.47 & 20.00 & 26.11 & 35.00 & 39.44 & 44.44 & 50.00 \\
\midrule
\textbf{ } & \textbf{Ours} & \textbf{7B} & \textbf{29.19} & \textbf{46.12} & \textbf{55.38} & \textbf{62.26} & \textbf{67.72} & \textbf{71.16} & \underline{36.63} & \textbf{50.99} & \textbf{59.90} & \textbf{68.81} & \textbf{73.27} & \textbf{79.21} \\
\midrule
\multicolumn{2}{l}{$\Delta$ vs Qwen3VL-235B-A22B-instruct} & - & \textbf{+12.06} & \textbf{+20.37} & \textbf{+23.42} & \textbf{+26.69} & \textbf{+27.94} & \textbf{+28.57} & \textbf{+1.2} & \textbf{+0.7} & \textbf{+0.7} & \textbf{+11.1} & \textbf{+12.7} & \textbf{+15.21} \\
\bottomrule
\end{tabular}
}
\end{table*}

\begin{table*}[htbp]
\centering
\caption{Ablation study of different components.}
\label{tab:ablation}
\resizebox{\textwidth}{!}{%
\begin{tabular}{l cccccc cccccc}
\toprule
\multirow{2}{*}{\textbf{Method}} & \multicolumn{6}{c}{\textbf{Fitzpatrick17k}} & \multicolumn{6}{c}{\textbf{Self-owned dataset}} \\
\cmidrule(lr){2-7} \cmidrule(lr){8-13}
 & TOP1 & TOP2 & TOP3 & TOP4 & TOP5 & TOP6 & TOP1 & TOP2 & TOP3 & TOP4 & TOP5 & TOP6 \\
\midrule
ours & \textbf{29.19} & \textbf{46.12} & \textbf{55.38} & \textbf{62.26} & \textbf{67.72} & \textbf{71.16} & \textbf{36.63} & 50.99 & 59.90 & \textbf{68.81} & \textbf{73.27} & \textbf{79.21} \\
ours(w/o DVE) & 24.45 & 33.24 & 42.22 & 48.62 & 54.15 & 57.69 & 35.64 & \textbf{51.49} & \textbf{60.89} & 66.83 & 70.79 & 74.75 \\
ours(w/o DVE \& stage 1) & 15.22 & 25.26 & 32.06 & 37.61 & 41.44 & 45.36 & 27.46 & 45.08 & 52.33 & 59.07 & 64.25 & 66.84 \\
\bottomrule
\end{tabular}%
}
\end{table*}

\begin{figure}[t]
  \includegraphics[width=\columnwidth]{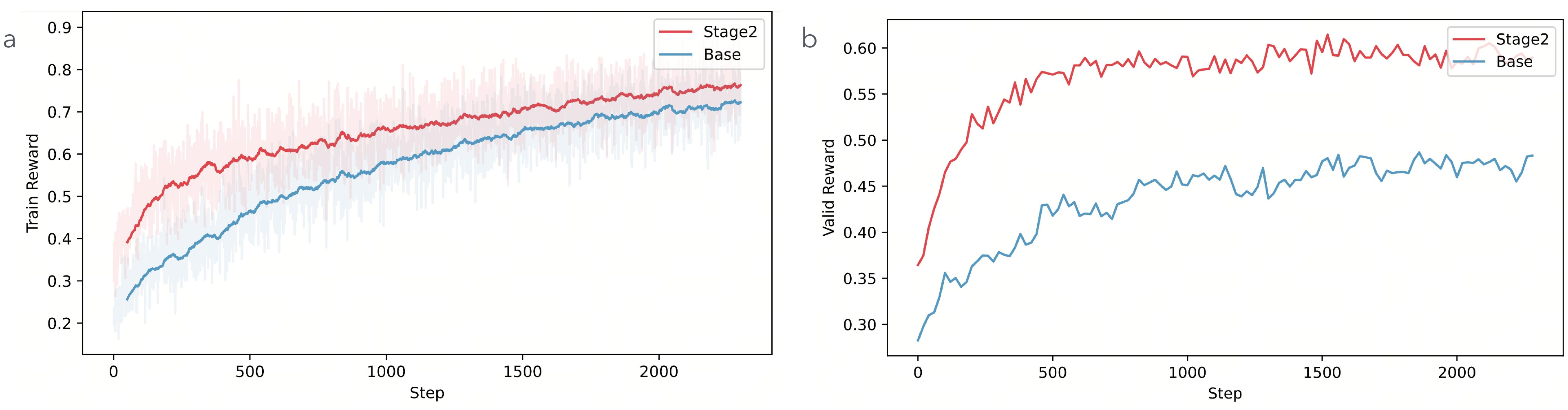}
  \caption{\textbf{Effectiveness of Stage 1 Caption Training}.
    (a) Training reward curves and (b) validation reward curves.
The blue line represents the model trained directly on the general-purpose baseline, while the red line denotes the model further trained based on the caption-enhanced Stage I model.}
  \label{fig:stage}
\end{figure}

\subsection{Ablation Studies and Analysis}
To investigate the individual contributions of the key components in our proposed method, specifically the Stage 1 captioning training task and the Dynamic Visual Encoding (DVE) module, we conducted a series of ablation experiments. The results are summarized in Table \ref{tab:ablation}. We compared three settings: (1) the baseline model without both DVE and the Stage 1 caption task, (2) the model with Stage 1 training but utilizing a static visual encoder (w/o DVE), and (3) our full model equipped with both components.

\vspace{1ex} 
\noindent\textbf{Effectiveness of Stage 1 Caption Training}
As shown in Figure \ref{fig:stage}, the Stage2 model exhibited faster convergence and achieved higher validation rewards compared with the Base model.
As shown in the last two rows of Table \ref{tab:ablation}, the introduction of the captioning task in Stage 1 yields a substantial performance improvement over the baseline. On the Self-owned dataset, the Top-1 accuracy surges from 27.46\% to 35.64\%, and on the Fitzpatrick17k dataset, it increases from 15.22\% to 24.45\%. This demonstrates that the caption generation task significantly enhances the model's ability to align visual features with textual descriptions, establishing a robust foundation for downstream disease identification.

\vspace{1ex} 
\noindent\textbf{Impact of Dynamic Visual Encoding (DVE)}
The incorporation of the DVE module further boosts performance, particularly in terms of generalization and top-ranked accuracy. Comparing the full model (ours) with the variant without DVE, we observe consistent improvements. On the Self-owned dataset, the Top-1 accuracy rises to 36.63\%, and the Top-6 accuracy reaches 79.21\%. Notably, the efficacy of DVE is most pronounced on the challenging Fitzpatrick17k dataset, where the Top-1 accuracy improves by approximately 4.74\% (from 24.45\% to 29.19\%), and the Top-6 accuracy sees a remarkable increase of over 13\% (from 57.69\% to 71.16\%). These results suggest that dynamic visual encoding allows the model to capture more fine-grained and adaptive visual features, which is critical for handling diverse skin conditions across different domains.

\vspace{1ex} 
\subsection{Qualitative Analysis: Visual Evidence Alignment}

\begin{figure*}[t]
\centering
\includegraphics[width=\linewidth]{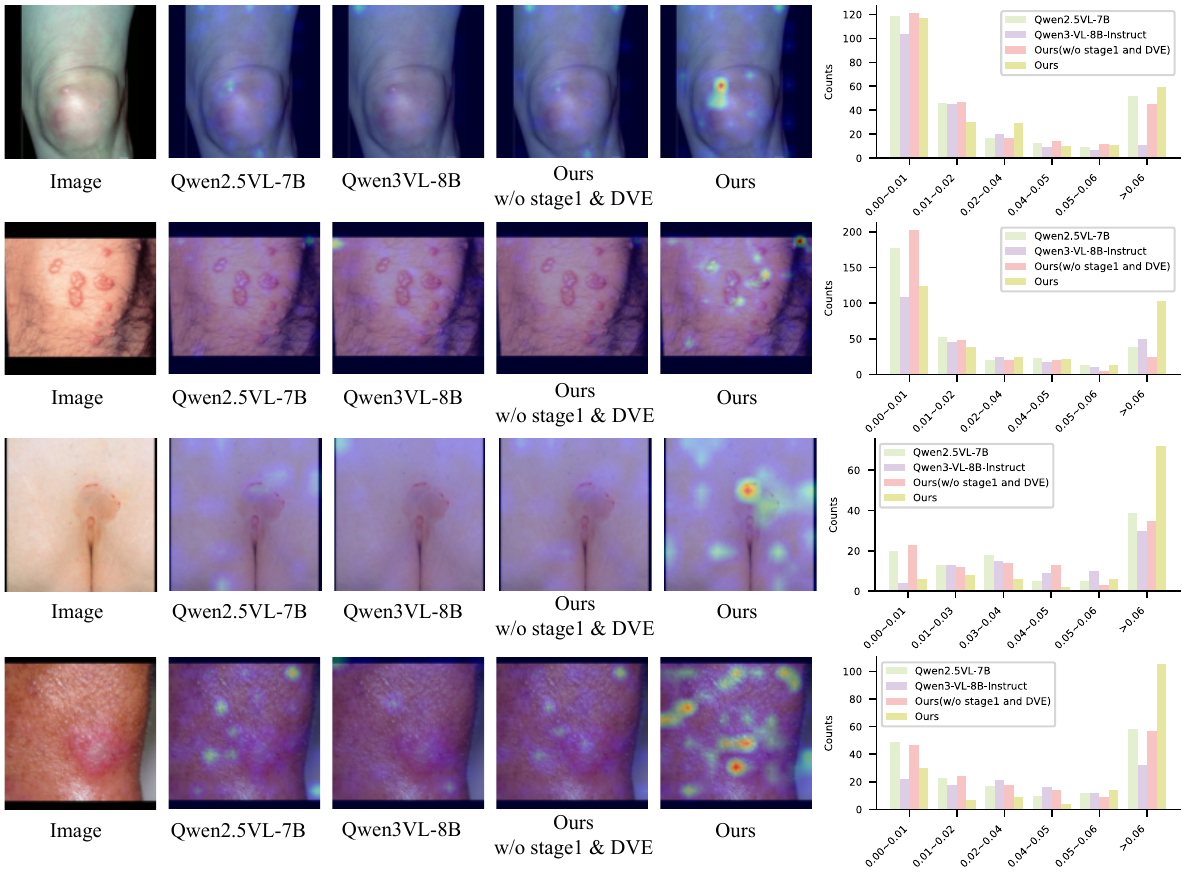} 
\caption{\textbf{Visual attention attribution analysis.} 
We visualize the cross-attention maps of the final diagnostic token relative to the input image across different models. The histograms (right) show the distribution of attention weights.
\textbf{(1) Attention concentration:} While baselines (Qwen2.5/3-VL) exhibit \textit{diffuse attention}, often distracted by healthy skin or background noise, Our method demonstrates precise \textit{lesion localization}, sharply focusing on pathological features (e.g., papules, ulcers).
\textbf{(2) Effect of FDLinear and stage 1 training:} Comparing "w/o stage1 \& DVE" with "Ours", the introduction of the Dynamic Vision Encoder (DVE) and stage 1 training significantly improve the model to adaptively highlights important diagnostic details.
\textbf{(3) Confidence shift:} The histograms reveal that our method (yellow) assigns higher attention weights ($>0.06$) to key regions than the baselines, indicating a shift from uncertain global scanning to confident diagnostic reasoning.}
\label{fig:attention_map}
\end{figure*}

To interpret the source of the proposed model's diagnostic superiority, we visualize the cross-attention maps corresponding to the final diagnostic token (Figure~\ref{fig:attention_map}). Three key phenomena observed in the heatmaps and histograms validate our architectural and training hypotheses:

\noindent\textbf{From global scanning to local fixation:} 
General-purpose MLLMs (Qwen2.5-VL, Qwen3-VL) tend to distribute attention broadly across the entire anatomical region (e.g., the whole knee in Row 1), exhibiting a "semantic spread" where the model fails to distinguish between the lesion and the surrounding healthy skin. In contrast, \textbf{our method} demonstrates distinct \textit{attention sparsity}, concentrating its focus exclusively on the pathological lesions. This confirms that the model has learned to ignore irrelevant background noise and lock onto the discriminative visual features essential for diagnosis.

\noindent\textbf{Enhanced disentanglement via DVE and stage 1:} 
Comparing the ablation variant ("w/o stage1 \& DVE") with the full pipeline, we observe that the complete method eliminates background artifacts more effectively. For instance, in complex scenarios with multiple scattered lesions (Row 2 and Row 4), the full model distinctly highlights separate lesion boundaries, whereas the ablation model produces a blurred or connected heatmap. This empirically supports our theoretical motivation: the Dynamic Vision Encoder (DVE), combined with the fine-grained visual guidance from stage 1, effectively "unfolds" the visual manifold. This allows the attention mechanism to linearly separate pathological textures from normal skin tones with high precision.

\noindent\textbf{High-confidence activation:} 
The attention weight distribution histograms (Figure~\ref{fig:attention_map}, right column) quantify this qualitative improvement. Baseline models follow a heavy-tailed distribution towards low weights ($0.00-0.02$), indicating a degree of uncertainty or "hesitation" in feature selection. Conversely, \textbf{our method} (yellow bars) exhibits a pronounced peak in the high-weight region ($>0.06$). This \textit{confidence shift} suggests that the proposed model performs diagnosis based on strong, localized visual evidence rather than weak global context.

\begin{figure}[t!]
  \centering
  \includegraphics[width=0.68\linewidth]{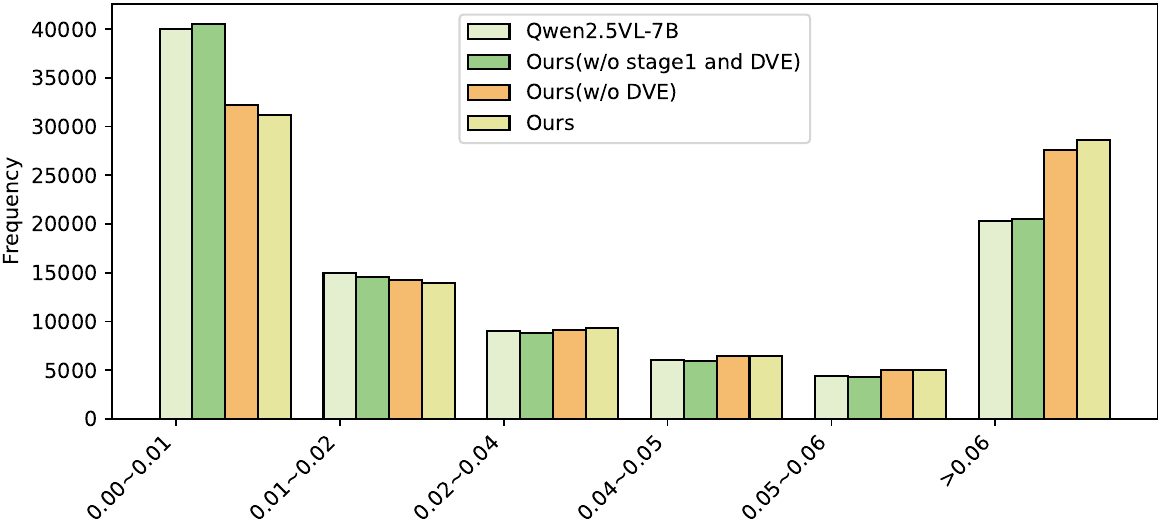} 
  \caption{\textbf{Quantitative distribution of attention weights across 500 test samples.} 
  The histogram statistics (scaled by 100) reveal a progressive shift in attention mechanisms. 
  \textbf{(1) Noise Suppression:} The full model (Ours, yellow) exhibits the lowest frequency in the background noise interval ($0.00 \sim 0.01$), substantially lower than the Qwen2.5-VL (light green).
  \textbf{(2) Signal Amplification:} In the high-confidence interval ($>0.06$), the introduction of Stage 1 (orange) drastically increases the frequency of focused attention. The integration of DVE (yellow) further boosts this peak, demonstrating that the full pipeline maximizes the signal-to-noise ratio in visual reasoning.}
  \label{fig:attn_stats}
\end{figure}

To verify that the observed attention concentration is a generalizable behavior rather than anecdotal evidence, we conducted a quantitative statistical analysis over 500 randomly sampled test images (Figure~\ref{fig:attn_stats}).
The distribution reveals a distinct "Shift-to-Right" phenomenon driven by our two-stage design:
\begin{itemize}
    \item \textbf{Baseline Uncertainty:} Qwen2.5-VL (light green) and the model without Stage 1 (dark green) show a heavy-tailed distribution dominated by the $0.00 \sim 0.01$ range. This indicates a "diffuse scanning" mode where probability mass is wasted on irrelevant background pixels.
    \item \textbf{Semantic Alignment via Stage 1:} Upon introducing caption learning (Orange, "w/o DVE"), we observe a sharp drop in the low-weight bin and a surge in the $>0.06$ bin. This confirms that medical captioning acts as a strong supervisor, teaching the model to distinguish pathological regions from healthy skin.
    \item \textbf{Feature Purification via DVE:} The full model (Yellow, "Ours") incorporating the Dynamic Vision Encoder achieves the most extreme distribution: it has the \textit{lowest} noise floor ($0.00 \sim 0.01$) and the \textit{highest} confidence peak ($>0.06$). This quantitatively proves that FDLinear effectively suppresses background redundancy and amplifying diagnostic signal strength.
\end{itemize}

\section{Limitations}

Model interpretability was not further evaluated in this study. After the second-stage training, we observed that the model tended to generate shorter captions as diagnostic evidence during end-to-end prediction. This behavior may affect the interpretability of its reasoning process. In future work, we plan to collaborate with dermatologists to design more systematic interpretability evaluation metrics and further refine the model accordingly.

All images used in this study were captured under relatively simple background conditions. Therefore, the model’s diagnostic performance may degrade when applied to real-world scenarios with complex or cluttered backgrounds. Future work will involve expanding the dataset to include more diverse imaging environments to improve robustness and generalization.

\section{Conclusion}

In this work, we addressed the fundamental limitations of general-purpose LVLMs in the domain of dermatological diagnosis. By framing the diagnostic process as an optimization of an image compression–decoding system, we demonstrated that the bottleneck in diagnostic performance stems from inefficient visual information transmission. Our proposed two-stage RL-based training strategy, coupled with the Dynamic Visual Encoding (DVE) module, effectively enables the model to disentangle critical pathological features from clinical background noise.

Our extensive experiments on both public datasets (Fitzpatrick17k) and self-owned datasets yield several key insights. First, we show that a domain-specialized 7B model can surpass massive models over 30$\times$ its size by enhancing visual-semantic alignment. Second, the proposed two-stage paradigm proves that explicit linguistic guidance (Stage I) and DVE module provides a necessary foundation for the reconstruction of implicit diagnostic cues (Stage II). Third, our qualitative attention analysis confirms a significant "confidence shift" from diffuse global scanning to localized diagnostic reasoning. Finally, by introducing a clinically grounded evaluation framework, we bridge the gap between algorithmic accuracy and real-world clinical actionability. 

We believe that this study provides a robust blueprint for developing high-efficiency, safety-aware medical AI. Future work will explore the generalization of this compression-decoding framework to other visually intensive medical specialties, such as pathology and radiology, to further validate the universality of our approach.

\bibliography{custom}

\appendix

\newpage
\section{Predefined Caption Schema}
\label{sec:appendix_pcs} 
\begin{tcolorbox}[left=0mm,right=0mm,top=0mm,bottom=0mm,boxsep=1mm,arc=0mm,boxrule=0pt, frame empty, breakable]

\begin{lstlisting}
{
    "color": "",                    // Detailed color-related information in the skin description; empty if none
    "location": "",                 // Description of all sites showing skin abnormality; empty if none
    "shape": "",                    // Information related to the described shape; empty if none
    "lesion_type": "",              // Type of skin lesion, e.g., macule, papule, nodule, vesicle, pustule, wheal; scales, erosion, ulcer, crust, lichenification; empty if none
    "number": "",                   // Count of abnormalities; empty if none
    "size": "",                     // Description of the abnormal skin's size; empty if none
    "texture": "",                  // Texture of the abnormal skin, e.g., hardness; empty if none
    "border_characteristics": "",   // Clarity of the abnormal skin's border; empty if none
    "surface_characteristics": "",  // All surface features, including central features; empty if none
    "distribution_characteristics": "", // Distribution features, e.g., scattered, isolated, dense, confluent; empty if none
    "surrounding_characteristics": "", // All features around the abnormal skin; empty if none
    "other": ""                     // Any other appearance descriptors that cannot fit into the fields above; only used when absolutely necessary, otherwise empty
}
\end{lstlisting}
\end{tcolorbox}

\section{Disease Diagnosis Prompt}
\label{sec:appendix_ddp} 
\begin{tcolorbox}[left=0mm,right=0mm,top=0mm,bottom=0mm,boxsep=1mm,arc=0mm,boxrule=0pt, frame empty, breakable]

\begin{lstlisting}
[Character]

As a dermatologist, please provide the most likely set of skin diseases (Top-{k}) based on the provided photos of skin diseases and explain the reasons. \n
[Output]
    First output the thinking process in <think> </think> tags and then output the final answer in <answer> </answer> tags.like this:
    <answer>{ """ + f'"top{flag[-1]}_diseases": '+ """[
        {
        "disease": "The name of the disease",
        "probability": 0.80,        // Between 0 and 1
        "key_matching_fields": ["color", "shape"], // Hit field
        "brief_reason": "A one-sentence reason "
        },
        ...
    ]
    }</answer>

\end{lstlisting}
\end{tcolorbox}

\section{Examples of disease diagnosis and examples of captions}
\label{sec:appendix_example}

\begin{figure}[htbp]
    \centering
    \includegraphics[width=0.68\linewidth]{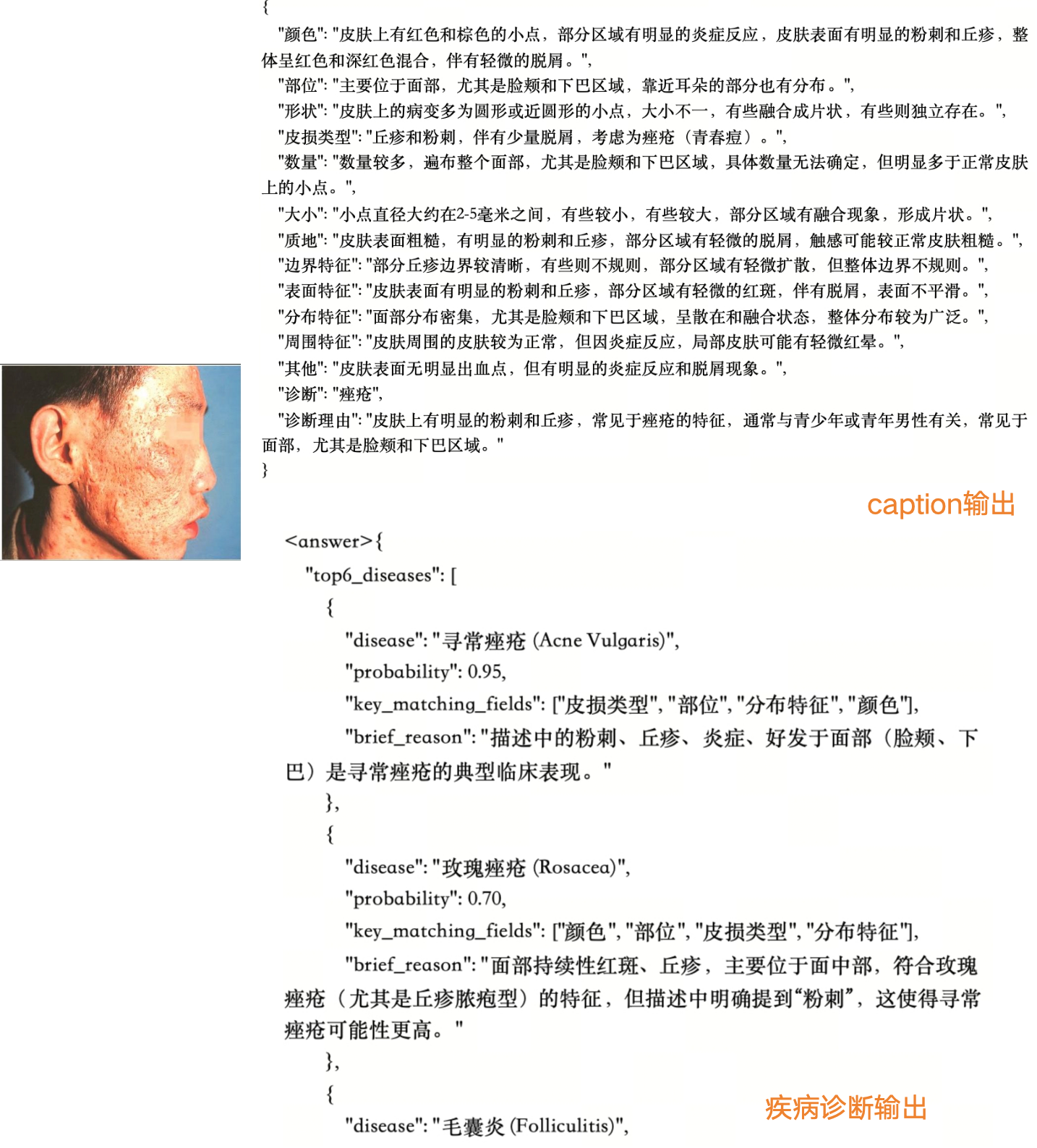} 
    \caption{\textbf{Examples of disease diagnosis and examples of captions.}} 
    \label{fig:demo}
\end{figure}

\end{CJK}
\end{document}